\documentclass[11pt]{article}

\usepackage{cite}
\usepackage{amsmath,amssymb,amsfonts}
\usepackage{algorithmic}
\usepackage{graphicx}
\usepackage{textcomp}
\usepackage{xcolor}

\usepackage[top=2.5cm, bottom=2.5cm, left=2cm, right=2cm]{geometry}
\newcommand{\argmin}{\operatornamewithlimits{argmin}}
\DeclareMathOperator{\sgn}{sgn}
\DeclareMathOperator{\Tr}{Tr}
\usepackage[ruled,vlined]{algorithm2e}
\usepackage{dsfont,bm}
\allowdisplaybreaks
\usepackage[colorlinks]{hyperref}
    
\usepackage{authblk}

\title{Robust PCA for Anomaly Detection and Data Imputation in Seasonal Time Series}
\author[1]{Hông-Lan Botterman}
\author[1]{Julien Roussel}
\author[1]{Thomas Morzadec}
\author[1]{Ali Jabbari}
\author[,1,2]{Nicolas Brunel\thanks{Corresponding author: \texttt{nicolas.brunel@ensiie.fr}}}

\affil[1]{Quantmetry, 52 rue d'Anjou, 75008 Paris, France}
\affil[2]{LaMME, ENSIIE, Université Paris Saclay, 1 square de la Résistance, 91025 Evry Cedex}

\date{}

\begin{document}

\maketitle

\begin{abstract}
We propose a robust principal component analysis (RPCA) framework to recover low-rank and sparse matrices from temporal observations. We develop an online version of the batch temporal algorithm in order to process larger datasets or streaming data. We empirically compare the proposed approaches with different RPCA frameworks and show their effectiveness in practical situations.
\end{abstract}

\noindent {\bf Keywords: } Robust principal component analysis, anomaly detection, data imputation, regularisation, adaptive estimation

\section{Introduction}
The ability to generate and collect data has increased considerably with the digitisation of our society. Unfortunately, this data is often corrupted or not collected in its entirety because of \textit{e.g.} faulty data collection systems, data privacy and so on. This is an issue for instance when tackling prediction tasks with a supervised learning algorithm. 

In this work, we are interested in anomaly detection and the subsequent imputation in time series. Our goal is to retrieve the smooth signal (plus possibly some white noise) from corrupted temporal data, as a preprocessing step for any downstream learning algorithm. Techniques to achieve these goals are extremely diverse and the interested reader can refer \textit{e.g.} to~\cite{blazquez2021review, shaukat2021review, Pratama2016, honaker2010missing} for a review or further details. Here, we reshape a noisy and corrupted time series into an appropriate matrix and we assume the smooth signal can be recovered since it corresponds to a low-rank matrix, while the anomalies are sparse.

Principal component analysis (PCA) is one of the most widely used techniques for dimension reduction in statistical data analysis. Given a collection of samples, PCA computes a linear projection of the samples to a low-dimensional subspace that minimises the $\ell_2$-projection error~\cite{eckart1936approximation, jolliffe2016principal}.
This method is proved effective in many real world applications, \textit{e.g.}~\cite{partridge1998fast, agarwal2010face, sanguansat2012principal}. However, its fragility in the face of missing or outlier data, common in everyday applications such as image processing or web data analysis, often puts its validity in jeopardy. In presence of a mixture of normal and abnormal measures, it is also difficult to select the dimension of the projection subspace representing the data.

To address this problem, a plethora of methods have been proposed for recovering low-rank and sparse matrices (aka robust principal component analysis (RPCA) \cite{wright2009robust}) with incomplete or grossly corrupted observations, such as principal component pursuit (PCP) \cite{candes2011robust} or outlier pursuit \cite{xu2010robust}. 
In principle, these methods aim at minimising an optimisation problem involving both trace norm and $\ell_1$-norm, convex relaxations for the rank function and the $\ell_0$-norm respectively \cite{candes2011robust}. Such relaxations bypass the fact that the basic problem is NP-hard.
RPCA is proving useful and effective in many applications such as video surveillance~\cite{luan2014extracting}, image and video processing~\cite{bouwmans2018applications}, speech recognition~\cite{gavrilescu2015noise} or latent segmentation indexing~\cite{deerwester1990indexing} just to name a few.

We are interested in anomaly detection in univariate time series and signal reconstruction: time series are first sized in matrices of appropriate dimensions (see Sec.~\ref{sec:results} for an example) in order to apply RPCA to detect anomalies and impute their values. Although many phenomena are modelled by time series and RPCA is a widely used tool, there is, to our knowledge, very little work specifically concerned with applying RPCA to time series. 
In~\cite{jin2017anomaly}, RPCA is specifically used to detect anomalies in time series of users' query records, but these anomalies are the result of a particular treatment of the sparse component. Hence, the problem to minimise resumes to the basic RPCA problem. 
In \cite{mardani2013recovery}, the authors introduce a compression matrix acting on the sparse one to reveal traffic anomalies. They show that it is possible to exactly recover the low rank and sparse components by solving a non-smooth convex optimisation problem. The work in~\cite{mardani2013robust} presents an estimator dedicated to recover sufficiently low-dimensional nominal traffic and sparse enough anomalies when the routing matrix is column-incoherent~\cite{candes2011robust}, and an adequate amount of flow counts are randomly sampled.
In~\cite{wang2018improved}, time is directly taken into account in the objective function: a constraint is added to maintain consistency between the rows of the low-rank matrix. The method is applied to subway passenger flow and is effective for detecting anomalies. However, to effectively recover anomalies, a filtering of the sparse matrix is performed, and the choice of the threshold value is debatable. Furthermore, their method is not directly applicable when there is missing or unobserved data.
In this paper, we aim to improve this RPCA approach, augmented by a consistency constraint between rows/columns, by taking into account missing data and noise, thus avoiding the debate of thresholding. 

It is worth mentioning that RPCA methods are parametric. These parameters drastically influence the results of the method, \textit{i.e.} the capacity of the algorithm to extract a smooth signal and to detect anomalies. 
For some formulations, there exist theoretical values that, in principle, allow the \textit{exact} recovery of the desired matrices~\cite{candes2009exact}. However, in practice and for real-world data, these values do not always work and it is often difficult to determine them without very precise knowledge of the data. We here seek to estimate all parameters from the examined data. More precisely, we propose to determine the parameters by cross-validation and consequently, we propose an end-to-end procedure to compute the RPCA. 

Batch algorithms suffer from some disadvantages: \textit{i}) they require a large amount of memory since all observed data must be stored in memory in order to be processed; \textit{ii}) they also become too slow to process large data; \textit{iii}) when data arrives continuously, they cannot simply adapt to this influx of information, but must restart from scratch and \textit{iv)} they do not adapt to data drift.
Henceforth, an online version of robust PCA method is highly necessary to process incremental data set, where the tracked subspace can be updated whenever new data is received.

Several RPCA methods have been developed but we can mainly group them into three categories/versions: Grassmannian Robust Adaptive Subspace Tracking Algorithm (GRASTA, \cite{he2011online}), Recursive Projected Compress Sensing (ReProCS, \cite{guo2014online, qiu2014recursive}) and Online Robust PCA via Stochastic Optimization (RPCA-STOC, \cite{feng2013online}).
GRASTA is built on Grassmannian Rank-One Update Subspace Estimation (GROUSE, \cite{balzano2010online}) and performs incremental gradient descent on the Grassmannian. 
ReProCS is mainly designed for solving problems for video sequence: separating a slowly changing background from moving foreground objects on-the-fly. 
A limitation of this approach is that it requires knowledge of the structure of the correlation model on the sparse part. Both GRASTA and Proc-ReProCS can only handle slowly changing subspaces. 
RPCA-STOC is based on stochastic optimisation for a close reformulation of the Robust PCA based on Principal Component Pursuit. In particular, RPCA-STOC splits the nuclear norm of the low-rank matrix as the sum of Frobenius norm of two matrices.

Lastly, note that RPCA-STOC works only with stable subspace, which could be a restriction for real applications, \textit{e.g.} data drift. In practice, at time $t$, RPCA-STOC updates the basis of subspace by minimising an empirical loss which involves all previously observed samples with equal weights.
In this work, we propose an online formulation RPCA with temporal regularisations by pursuing the idea of RPCA-STOC. 
We adapt the algorithm to cope with unstable subspaces by considering an online moving window, \textit{i.e.} by considering an empirical loss based only on the most recent samples.

The rest of the paper is organised as follows. In Sec.~\ref{sec:background} we recall some notions and present the different formulations of RPCA and deep dive into the algorithms in Sec.~\ref{sec:method}. In Sec.~\ref{sec:results}, we compare the different methods on synthetic and real-world data and we finally conclude in Sec.~\ref{sec:discussion}.

\section{Background and Problem Formulation}
\label{sec:background}

Let $\mathbf{y} = \{y_i\}_{1 \leq i \leq I}$ be a univariate time series, where indices correspond to uniform time steps. Let $T_0$ be the main seasonality (or period) of this time series. We assume that we observe $n \in \mathbb{N}^+$ periods in $\mathbf{y}$, such that $I = T_0 \times n$; otherwise we add missing values at the end of the signal to fit in. We consider the case where the time series $\mathbf{y}$ is a corrupted version of a true signal: the corruptions can be multiple such as missing data, presence of anomalies, and additive observation noise. We propose to filter and reconstruct the whole signal by using the similarity of $n$ different periods of length $T_0$. 
For this reason, we convert $\mathbf{y}$ into a matrix $\mathbf{D} \in \mathbb{R}^{T_0 \times n}$ such that $\mathbf{D} = [\mathbf{d}_1 \vert \dots \vert \mathbf{d}_n]$ with $\mathbf{d}_t = [y_{(t-1)T_0+1}, \cdots, y_{(t-1)T_0+T_0}]^\top, \, t=1,\cdots,n$. We consider that the periodicity of $\mathbf{y}$ and the repetition of patterns across time means that the different periods $\mathbf{d}_t, t=1,\cdots,n$ lie in a vector space of small dimension $r$. In favorable situations, this subspace can be estimated with a PCA of the matrix $\mathbf{D}$ but the assumed corruption of the data requires to use a robust version. In addition, a noticeable change with respect to standard PCA or RPCA is that the observations $\mathbf{d}_t, \mathbf{d}_{t+T_k}$ are not independent: we show how we can take advantage of the similarity between $\mathbf{d}_t$ and $\mathbf{d}_{t+T_k}$ for improving the global reconstruction of the signal.

In the following, $\Vert \mathbf{M} \Vert_* = \sum_{\sigma \in \mathrm{Sp}(\mathbf{M})} \sigma$ is the nuclear norm of $\mathbf{M}$; $\Vert \mathbf{M} \Vert_1 = \sum_{ij} \vert M_{ij} \vert$ is the $\ell_1$-norm of $\mathbf{M}$ seen as a vector and $\Vert \mathbf{M} \Vert_F^2 = \sum_{ij} M_{ij}^2$. The $\Omega$ set is the set of observed data and  $\mathcal{P}_{\Omega}(\mathbf{M})$ is the projection of $\mathbf{M}$ on $\Omega$, \textit{i.e.} $\mathcal{P}_{\Omega}(\mathbf{M})_{ij} = M_{ij}$ if $(i,j) \in \Omega$, 0 otherwise.

\subsection{Robust Principal Component Analysis}
Suppose we observe some data $\mathbf{y}$ or equivalently some entries of an incomplete, noisy and corrupted matrix $\mathbf{D}$. We seek to compute the following decomposition $\mathbf{D} = \mathbf{X} + \mathbf{A} + \mathbf{E}$, where $\mathbf{X}$ is a low-rank matrix whose rank is upper bounded by $r \leq \min \{T_0, n \}$, $\mathbf{A}$ is the sparse matrix of anomalies and $\mathbf{E}$ represents a Gaussian random fluctuation. In addition, we assume that for whatever reason, we only observe a subset of entries $\Omega \subseteq \{1, \cdots, T_0\} \times \{1, \cdots, n\}$. 
The RPCA formulation is as follows
\begin{equation}
    \min_{\substack{ \mathbf{X}, \mathbf{\tilde{A}} \\ \text{ s.t. } \mathcal{P}_{\Omega} (\mathbf{D}) =  \mathcal{P}_{\Omega} (\mathbf{X} + \mathbf{\tilde{A}}) }} \Vert \mathbf{X} \Vert_* + \lambda_2 \Vert \mathbf{\tilde{A}} \Vert_1,
    \label{eq:RPCA}
\end{equation}
where $\lambda_2 > 0$ is the inverse sensitivity parameter. Here we threshold out the \textit{significant} anomalies $\mathbf{A}$ from the noise $\mathbf{E}$, based on the anomaly score $s_{ij} = \vert \tilde{A}_{ij} \vert / \sigma_{X_{i:}}$: given a threshold $\alpha>0$, $\mathbf{\tilde{A}} = \mathbf{A} + \mathbf{E}$ with $A_{ij} = \tilde{A}_{ij} \mathds{1}_{s_{ij} > \alpha}$ and $E_{ij} = \tilde{A}_{ij} \mathds{1}_{s_{ij} \leq \alpha }$. Here $\sigma_{X_{i:}}$ denotes the standard deviation of the row $i$. This sparsifying post-processing on the  matrix $\mathbf{\tilde{A}}$ is quite common, see for instance~\cite{wang2018improved}.

This separation only makes sense if the low-rank part $\mathbf{X}$ is not sparse. This is related to the notion of \textit{incoherence} for matrix completion problem introduced in~\cite{candes2009exact} which imposes conditions on the singular vectors of the low-rank matrix. There are also issues when $\mathbf{\tilde{A}}$ is low-rank, in particular when data is missing not at random. Under these conditions, the main result of \cite{candes2011robust} states it is possible to \textit{exactly} recover the low-rank and sparse components.

\subsection{Robust Principal Component Analysis with Temporal Regularisations}
In the case of independent observations, we only have to take into account the correlations between variables, whereas time series exhibits a dependency structure. When the time series has a seasonality of periodicity $T_0$, we expect the correctly sized matrix to have low rank and adjacent columns to be similar (see Sec.~\ref{sec:results}).
However, to the best of our knowledge, not much specific temporal or time series oriented RPCA has been developed yet~\cite{mardani2013recovery, mardani2013robust, wang2018improved}. 

Here we use a set of $K$ terms penalising the differences between columns spaced by given times $T_k$: $\Vert \mathbf{XH_k} \Vert_F^2$, where the $\{ \mathbf{H_k} \}_{1 \leq k \leq K}$ are Toeplitz matrices with 1 on the diagonal, -1 on the super-$T_k$ diagonal and 0 otherwise. 
Stated otherwise, $T_k$ is the number of columns between two supposedly similar ones. A typical example is to use $\mathbf{H_1}$ for controlling the similarity between $\mathbf{d}_t$ and $\mathbf{d}_{t+1}$, \textit{i.e.} $T_k=1$. Another example: if each column of a matrix represents a day, we can put $T_k=7$ since we expect each day of the week to be``similar'', \textit{i.e.} similarity between $\mathbf{d}_t$ and $\mathbf{d}_{t+7}$.
In addition and in order to explicitly take noise into account, we relax the constraint $\mathbf{D} = \mathbf{X} + \mathbf{\tilde{A}}$ to eventually get the following problem 
\begin{align}
        \min_{\mathbf{X}, \mathbf{A}} 
        \frac{1}{2} \Vert \mathcal{P}_{\Omega}(\mathbf{D} - \mathbf{X} -\mathbf{A}) \Vert_F^2
        + \lambda_1  \Vert \mathbf{X} \Vert_* 
        + \lambda_2 \Vert \mathbf{A} \Vert_1  + \sum_{k=1}^K \eta_k \Vert \mathbf{XH_k} 
        \Vert_F^2
    \label{eq:RPCA_temporal}
\end{align}
where $\lambda_1, \lambda_2$ and $\{\eta_k\}_k$ are real parameters. 

Note that eq.~\eqref{eq:RPCA_temporal} can be related to RPCA on graph (see \textit{e.g.}~\cite{shahid2015robust}), where nodes have at most one outgoing link (our graph is a chain). Indeed, each $\mathbf{H_k}$ can be associated with an adjacency matrix $\mathbf{\hat{H}_k}$ defined as
\begin{equation*}
   \mathbf{\hat{H}_k} := \mathbf{I_n} - 
\left[
\begin{array}{c|c}
  \mathbf{I_{T_k}} & \raisebox{-15pt}{{\large \mbox{{$ \quad \mathbf{H_k} \quad$}}}} \\[-4ex]
  \phantom{\vdots} & \\[-0.5ex]
  \mathbf{0} &
\end{array}
\right],
\end{equation*}
with $\mathbf{I_a}$ the identity matrix of dimension $a$. By defining the Laplacian associated with this graph $\mathbf{\mathcal{L}}_k = \mathbf{Dg} - \mathbf{\hat{H}_k}$ ($\mathbf{Dg}$ is the degree matrix), we find a similar formulation to \cite{shahid2015robust} of the regularisation, \textit{i.e.}  $\Vert \mathbf{XH_k} \Vert_F^2 = \Tr(\mathbf{X}^\top\mathbf{\mathcal{L}}_k\mathbf{X})$. However, contrary to their problem, we force the matrix $\mathbf{X}$ to be low-rank.
Hence $\mathbf{H_k}$ matrices select an appropriate ``neighbourhood” for each observation and measure similarity, and the choice of $\mathbf{H_k}$ offers an avenue for modelling complex dependencies.

One can solve this problem (eq.~\eqref{eq:RPCA_temporal}) by computing its associated augmented Lagrangian and using the Alternating Direction Method of Multipliers~\cite{boyd2011distributed}. 

\subsection{Online Robust Principal Component Analysis with Temporal Regularisations}
We now turn to the online formulation of the problem~\eqref{eq:RPCA_temporal}. For this, recall the nuclear norm of a matrix $\mathbf{X} \in \mathbb{R}^{T_0 \times n}$ whose rank is upper bounded by $r$ can be expressed as 
\begin{equation}
    \Vert \mathbf{X} \Vert_* 
    = \inf_{\substack{ \mathbf{L} \in \mathbb{R}^{T_0 \times r} \\ \mathbf{Q} \in \mathbb{R}^{n \times r}}} 
    \left\{ 
        \frac{1}{2} ( \Vert \mathbf{L} \Vert_F^2 +  \Vert \mathbf{Q} \Vert_F^2 ) : \mathbf{X} = \mathbf{LQ^\top} 
    \right\}. 
\end{equation}
$\mathbf{L}$ can be interpreted as the basis for low-rank subspace while $\mathbf{Q}$ represents the coefficients of the observations with respect to that basis.
Plugging this expression into eq.~\eqref{eq:RPCA_temporal}, the problem becomes
\begin{align}
    \begin{split}
    \min_{\substack{\mathbf{X}, \mathbf{A}, \mathbf{L}, \mathbf{Q} \\ \text{s.t. } \mathbf{X} = \mathbf{LQ}^\top}} \, 
    & \frac{1}{2} \Vert \mathcal{P}_{\Omega}( \mathbf{D} - \mathbf{X} - \mathbf{A}) \Vert_F^2 
    + \frac{\lambda_1}{2} \left( \Vert \mathbf{L} \Vert_F^2 + \Vert \mathbf{Q} \Vert_F^2 \right) 
    + \lambda_2 \Vert \mathbf{A} \Vert_1 
    + \sum_{k=1}^K \eta_k \Vert \mathbf{XH_k} \Vert_F^2.
    \end{split}
\end{align}

Given a finite set of samples $\mathbf{D} = [\mathbf{d}_1, . . . , \mathbf{d}_n] \in \mathbb{R}^{T_0 \times n}$, solving problem~\eqref{eq:RPCA_temporal} amounts to minimising the following empirical cost function
\begin{equation}
 f_n(\mathbf{L}) := \frac{1}{n} \sum_{i=1}^n l(\mathbf{d}_i, \mathbf{L}) + \frac{\lambda_1}{2n} \Vert \mathbf{L} \Vert_F^2
\end{equation}
where  the loss function for each sample is defined as
\begin{align}
\begin{split}
 l(\mathbf{d}_i, \mathbf{L}) 
 & := \min_{\mathbf{q}, \mathbf{a}} h(\mathbf{d}_i, \mathbf{L}, \mathbf{q}, \mathbf{a}) \\
 & := \min_{\mathbf{q}, \mathbf{a}} \frac{1}{2} \Vert \mathcal{P}_{\Omega}(  \mathbf{d}_i - \mathbf{Lq} - \mathbf{a} ) \Vert_2^2 + \frac{\lambda_1}{2} \Vert \mathbf{q} \Vert_2^2 + \lambda_2 \Vert \mathbf{a} \Vert_1  
 + \sum_{k=1}^K \eta_k \Vert \mathbf{Lq} - \mathbf{Lq}_{-T_k} \Vert_2^2,
\end{split}
\end{align}
with $\mathbf{q}_{-T_k}$ the $T_k$-th last computed vector $\mathbf{q}$.

In stochastic optimisation, one is usually interested in minimising the expected cost over all the samples, \textit{i.e.} $f(\mathbf{L}) := \mathbb{E}_\mathbf{d} [ l(\mathbf{d}, \mathbf{L}) ] = \lim_{n \rightarrow \infty} f_n(\mathbf{L})$ where the expectation is taken with respect to the distribution of the samples $\mathbf{d}$.

To solve this problem, we borrow ideas from~\cite{feng2013online} that consist in optimising the coefficients $\mathbf{q}$, anomalies $\mathbf{a}$ and basis $\mathbf{L}$ in an alternative manner, \textit{i.e.} solving one vector while keeping the others fixed. In the $t$-th time instance ($0 \leq t \leq n$), $\mathbf{q}_t$ and $\mathbf{a}_t$ are solutions of
\begin{align}
    \begin{split}
    \{\mathbf{q}_t, \mathbf{a}_t \} = 
    & \argmin_{\mathbf{q}, \mathbf{a}} \frac{1}{2} \Vert \mathbf{d}_t - \mathbf{L}_{t-1}\mathbf{q} - \mathbf{a} \Vert_2^2 + \frac{\lambda_1}{2} \Vert \mathbf{q} \Vert_2^2 + \lambda_2 \Vert \mathbf{a} \Vert_1 
    + \sum_{k=1}^K \eta_k \Vert \mathbf{L}_{t-1}\mathbf{q} - \mathbf{L}_{t-1}\mathbf{q}_{-T_k} \Vert_2^2.
    \end{split}
\end{align}
We then obtain the estimation of the basis $\mathbf{L}_t$ through minimising the cumulative loss with respect to the previously estimated coefficients $\{\mathbf{q}_i\}_{i=1}^t$ and anomalies $\{\mathbf{a}_i\}_{i=1}^t$. To do so, we define a surrogate function (and an upper bound) of the empirical cost function $f_t(\mathbf{L})$ as
\begin{align}
    \begin{split}
    g_t(\mathbf{L}) 
    & := \frac{1}{t} \sum_{i=1}^t h(\mathbf{d}_i, \mathbf{L}, \mathbf{q}_i, \mathbf{a}_i) + \frac{\lambda_1}{2t} \Vert \mathbf{L} \Vert_F^2 \\
    & \geq \frac{1}{t} \sum_{i=1}^t \min_{\mathbf{q}, \mathbf{a}} h(\mathbf{d}_i, \mathbf{L}, \mathbf{q}, \mathbf{a}) + \frac{\lambda_1}{2t} \Vert \mathbf{L} \Vert_F^2 = f_t(\mathbf{L}).
    \end{split}
    \label{eq:online_glt}
\end{align}
Minimising this function yields the estimation of $\mathbf{L}_t$; the solution is given by
\begin{align}
    \begin{split}
    \mathbf{L}_t  = 
    & \left[ \sum_{i=1}^t (\mathbf{d}_i - \mathbf{a}_i) \mathbf{q}_i^\top \right] \left[ \sum_{i=1}^t \mathbf{q}_i\mathbf{q}_i^\top + 2 \sum_{k=1}^K \eta_k (\mathbf{q}_t - \mathbf{q}_{t-T_k}) (\mathbf{q}_t - \mathbf{q}_{t-T_k})^\top + \lambda_1 \mathbf{I} \right]^{-1}
    \end{split}
\end{align}
In practice $\mathbf{L}_t$ can be quickly updated by block-coordinate descent with warm restarts~\cite{loshchilov2016sgdr}.

One limitation of this method is that it assumes a stable subspace. As it can be seen in applications, for instance background substraction in video surveillance or anomaly detection in financial datasets, the stability of the signal subspace is a restrictive assumption that needs to be relaxed. Our eq.~\eqref{eq:online_glt} makes this assumption because we seek to minimise the function considering all the past (from 1 to $t$), each sample having the same importance. This is obviously undesirable if the underlying subspace changes over time. To avoid such a problem, we combine common idea of time series--namely moving windows--with stochastic RPCA with temporal regularisations. 
Concretely, we update the basis $\mathbf{L}_t$ from the last $n_w$ (\textit{i.e.} most recent) samples. This translates into a slight modification of eq.~\eqref{eq:online_glt} as follow
\begin{align}
    \begin{split}
    g^w_t (\mathbf{L}) := 
    & \frac{1}{n_w} \sum_{i=t-n_w}^t \left( \frac{1}{2} \Vert \mathcal{P}_{\Omega}(  \mathbf{d}_i - \mathbf{Lq}_i - \mathbf{a}_i ) \Vert_2^2 + \frac{\lambda_1}{2} \Vert \mathbf{q}_i \Vert_2^2 \right. \\
    & \left.  + \lambda_2 \Vert \mathbf{a}_i \Vert_1 + \sum_{k=1}^K \eta_k \Vert \mathbf{Lq}_i - \mathbf{Lq}_{i-T_k} \Vert_2^2 \right) + \frac{\lambda_1}{2n_w} \Vert \mathbf{L} \Vert_F^2.
    \end{split}
    \label{eq:online_gltw},
\end{align}
This formulation has the ability to \textit{quickly} adapt to changes in the underlying subspace.

\section{Method}
\label{sec:method}

We present the algorithms for the batch and online RPCA with temporal regularisations. To simplify notation, the period $T_0$ (time) is $m$ in the algorithms.

\subsection{Batch Temporal algorithm}
The associated augmented Lagrangian of eq.~\eqref{eq:RPCA_temporal} is 
\begin{align}
        \mathcal{L}_{\mu} & (\mathbf{X}, \mathbf{L}, \mathbf{Q}, \mathbf{A}, \{\mathbf{R_k}\}_{k = 1}^{K}, \mathbf{Y}) = 
        \frac{1}{2} \Vert \mathcal{P}_{\Omega} ( \mathbf{X} + \mathbf{A} - \mathbf{D} ) \Vert_F^2
        + \frac{\lambda_1}{2} \Vert \mathbf{L} \Vert_F^2
        + \frac{\lambda_1}{2} \Vert \mathbf{Q} \Vert_F^2 \nonumber \\ 
        & + \lambda_2 \Vert \mathbf{A} \Vert_1
        + \sum_{k=1}^K \eta_k \Vert \mathbf{R_k} \Vert_F^2
         + \langle \mathbf{Y}, \mathbf{X} - \mathbf{LQ}^\top \rangle 
         + \frac{\mu}{2} \Vert \mathbf{X} - \mathbf{LQ}^\top \Vert_F^2  
    \label{eq:lagrangian}
\end{align}

The Alternating Direction Method of Multipliers (ADMM, \cite{boyd2011distributed}) has proven to be efficient in solving this type of problems~\cite{candes2011robust, wang2018improved}. More precisely, by using the fact that $2 \langle \mathbf{a},\mathbf{b} \rangle + \Vert \mathbf{b} \Vert^2_F = \Vert \mathbf{a} + \mathbf{b} \Vert^2_F -  \Vert \mathbf{a} \Vert^2_F$ and by zeroing the derivatives, it is straightforward to obtain the closed-form solutions. 
Moreover, recall that $\mathcal{S}_{\alpha}(\mathbf{Z}) := \sgn(\mathbf{Z}) \max (\vert \mathbf{Z} \vert - \alpha, 0 )$ is the soft thresholding operator at level $\alpha$, applied element-wise and is the solution of $\nabla_{\mathbf{B}} (\Vert \mathbf{Z} - \mathbf{B}) \Vert_F^2  + \alpha \Vert \mathbf{B} \Vert_1) = 0.$
All steps are summarised in Algorithm~\ref{algo:batch_RPCA}. Since min($m,n$) $>r$, time complexity at each iteration is $\mathcal{O}(m^2n)$ while space complexity equals $\mathcal{O}(mn)$.

\begin{algorithm}
\SetAlgoLined
\KwData{observations $\mathbf{D} \in \mathbb{R}^{m \times n}$, params $\lambda_1, \lambda_2, \{\eta_k\}_{1 \leq k \leq K}$}
\KwResult{low-rank matrix $\mathbf{X} \in \mathbb{R}^{m \times n}$, sparse matrix $\mathbf{A} \in \mathbb{R}^{m \times n}$}
 initialisation: $\mathbf{X}^{(0)} = \mathbf{A}^{(0)} = \mathbf{Y}^{(0)} = \mathbf{1} \in \mathbb{R}^{m \times n}$, $\mathbf{L}^{(0)} = \mathbf{1} \in \mathbb{R}^{m \times r}$, $\mathbf{Q}^{(0)} = \mathbf{1} \in \mathbb{R}^{n \times r}$, $\mu$ = 1e-6, $\mu_{max}$ = 1e10, $\rho$ = 1.1, $\epsilon$ = 1e-8, max\_iter = 1e6, $i=0$ \;
 \While{ $e > \epsilon$ and i $<$ max\_iter}{
    $\mathbf{X}^{(i+1)} \leftarrow 
        \left(
            \mathbf{D} - \mathbf{A}^{(i)} + \mu \mathbf{L}^{(i)}\mathbf{Q}^{(i)\top} - \mathbf{Y}
        \right)
        \left(
            (1+\mu)\mathbf{I} + 2 \sum\nolimits_k \eta_k \mathbf{H_k}\mathbf{H_k}^\top
        \right)^{-1}$  
    \;
    $\mathbf{A}^{(i+1)} \leftarrow 
        \mathcal{S}_{\lambda_2 / 2} (\mathbf{D} - \mathbf{X}^{(i+1)}) 
    $  \;
    $\mathbf{L}^{(i+1)} \leftarrow 
        \left(
            \mu \mathbf{X}^{(i+1)} + \mathbf{Y}^{(i)}
        \right) 
        \mathbf{Q}^{(i)} 
        \left(
            \lambda_1 \mathbf{I} + \mu \mathbf{Q}^{(i)^\top} \mathbf{Q}^{(i)} 
        \right)^{-1} 
    $  \;
    $\mathbf{Q}^{(i+1)} \leftarrow 
        \left(
            \mu \mathbf{X}^{(i+1)} + \mathbf{Y}^{(i)} 
        \right)^\top
        \mathbf{L}^{(i+1)} 
        \left(
            \lambda_1 \mathbf{I} + \mu \mathbf{L}^{(i+1)\top}\mathbf{L}^{(i+1)} 
        \right)^{-1} 
    $  \;
    $\mathbf{Y}^{(i+1)} \leftarrow 
        \mathbf{Y}^{(i)} + \mu (\mathbf{X}^{(i+1)} - \mathbf{L}^{(i+1)} \mathbf{Q}^{(i+1)^\top})
    $  \;    
    $e = \max \left\{ \Vert \mathbf{M}^{(t+1)} - \mathbf{M}^{(t)} \Vert_{\infty}, \, \mathbf{M} \in \{\mathbf{X}, \mathbf{A}, \mathbf{L}, \mathbf{Q} \} \right\} $ \;
    $\mu \leftarrow \min(\rho \mu, \mu_{max})$ \;
    $i \leftarrow i+1$
 }
\caption{Batch Temporal Robust Principal Component Analysis}
\label{algo:batch_RPCA}
\end{algorithm}

\subsection{Online Temporal algorithm via stochastic optimisation}
We here present the algorithm for solving the online RPCA with temporal regularisations (Algorithm~\ref{algo:stochasticOptimisation}).
First, we compute the batch algorithm on the $n_{burnin}$ first columns of $\mathbf{D}$, \textit{i.e.} the columns we already observed. Then, we alternatively update the coefficients, the anomalies and the basis with off-the-shelf convex optimisation solver (Algorithm~\ref{algo:projection}) and block-coordinate descent with warm restarts (Algorithm~\ref{algo:basisUpdate}) respectively.
Here, time complexity at each iteration of the online part is $\mathcal{O}(mr^2)$ (and for the batch part on the $n_{burnin}$ samples: $\mathcal{O}(m^2n_{burnin})$) while space complexity equals $\mathcal{O}(mr)$ (and $\mathcal{O}(mn_{burnin})$ for the batch part). When processing \textit{big data}--especially when $n\gg m>r$--there is a computational complexity advantage to using the online algorithm since we will have $\frac{r^2}{mB} < 1$ with $B$ the number of iterations of the temporal batch algorithm.

\begin{algorithm}
\KwData{observations $\mathbf{D} = [ \mathbf{d}_1, \mathbf{d}_2, ..., \mathbf{d}_n] \in \mathbb{R}^{m \times n}$ (which are revealed sequentially), 
params $\lambda_1, \lambda_2, \{\eta_k\}_{1 \leq k \leq K}$,
$\mathbf{D}_b = [ \mathbf{d}_{-n_{burnin}}, ..., \mathbf{d}_{-1}] \in \mathbb{R}^{m \times n_{burnin}}$
}
\KwResult{low-rank matrix $\mathbf{X} \in \mathbb{R}^{m \times n}$, 
sparse matrix $\mathbf{A} \in \mathbb{R}^{m \times n}$
}
 initialisation: $\mathbf{X}^{(0)} = \mathbf{A}^{(0)} = \mathbf{Y_0}^{(0)} = \mathbf{1} \in \mathbb{R}^{m \times T_0}$, $\mathbf{L}^{(0)} = \mathbf{1} \in \mathbb{R}^{m \times r}$, $\mathbf{Q}^{(0)} = \mathbf{1} \in \mathbb{R}^{T_0 \times r}$, $\mu$ = 1e-6, $\mu_{max}$ = 1e10, $\rho$ = 1.1, $\epsilon$ = 1e-8, max\_iter = 1e6, $k=0$, $n_{burnin}$, $n_w$ \;
 Compute RPCA on burn in samples via Algorithm \ref{algo:batch_RPCA}: \\
 $\quad \mathbf{X}_{b}, \mathbf{A}_{b}$ = batch\_RPCA($\mathbf{D}_b$) \;
 Compute SVD decomposition of $\mathbf{X}_{b}$: \\
 $\quad \mathbf{X}_{b} = \mathbf{U}\mathbf{\Sigma}\mathbf{V}^\top$ \;
 Initialise $\mathbf{L}$, $\mathbf{B}$ and $\mathbf{C}$: \\
    $\quad \mathbf{L}_0 \leftarrow \mathbf{U}\mathbf{\Sigma}^{1/2}$ \\
    $\quad \mathbf{B}_0 \leftarrow \sum_{i=1}^{n_{burnin}} \mathbf{q}_i\mathbf{q}_i^\top + 2 \sum_{k=1, T_k>i}^K  \eta_k (\mathbf{q}_i - \mathbf{q}_{i-T_k}) (\mathbf{q}_t - \mathbf{q}_{i-T_k})^\top$ \\
    $\quad \mathbf{C}_0 \leftarrow \sum_{i=1}^{n_{burnin}} (\mathbf{q}_i - \mathbf{a}_i) \mathbf{q}_i^\top$ \;
 \For{ $t = 1$ to $n$ }{
    Reveal sample $\mathbf{d}_t$ \;
    Project this new sample via Algorithm~\ref{algo:projection}: \\
        $\{\mathbf{q}_t, \mathbf{a}_t \} = \argmin_{\mathbf{q}, \mathbf{a}} \frac{1}{2} \Vert \mathbf{d}_t - \mathbf{L}_{t-1}\mathbf{q} - \mathbf{a} \Vert_2^2 + \frac{\lambda_1}{2} \Vert \mathbf{q} \Vert_2^2 + \lambda_2 \Vert \mathbf{a} \Vert_1 + \sum_{k=1}^K \eta_k \Vert \mathbf{L}_{t-1}\mathbf{q} - \mathbf{L}_{t-1}\mathbf{q}_{-T_k} \Vert_2^2$ \;
    Update $\mathbf{B}$ and $\mathbf{C}$: \\
        $\quad \mathbf{B}_t \leftarrow \mathbf{B}_{t-1} + \mathbf{q}_t\mathbf{q}_t^\top + 2 \sum_{k=1}^K  \eta_k (\mathbf{q}_t - \mathbf{q}_{t-T_k}) (\mathbf{q}_t - \mathbf{q}_{t-T_k})^\top$ - $\mathbf{q}_{t-n_w}\mathbf{q}_{t-n_w}^\top$\\
        $\quad \mathbf{C}_t \leftarrow \mathbf{C}_{t-1} + (\mathbf{d}_t - \mathbf{a}_t) \mathbf{q}_t^\top$ - $(\mathbf{d}_{t-n_w}-\mathbf{a}_{t-n_w})\mathbf{q}_{t-n_w}^\top$ \;
    Compute $\mathbf{L}_t$ with $\mathbf{L}_{t-1}$ as warm restart using Algorithm \ref{algo:basisUpdate}: \\
        $\quad \mathbf{L}_t = \argmin_{\mathbf{L}} \Tr (\mathbf{L}^\top (\mathbf{B}_t + \lambda_1 \mathbf{I}) \mathbf{L} ) - \Tr (\mathbf{L}^\top \mathbf{C}_t)$
 }
 Return $\mathbf{X} = \mathbf{L}_n\mathbf{Q}^\top$ (low-rank matrix, $\mathbf{Q}$ is stacked $\mathbf{q}$), $\mathbf{A}$ (sparse matrix, $\mathbf{A}$ is stacked $\mathbf{a}$)
 \caption{Stochastic optimisation}
 \label{algo:stochasticOptimisation}
\end{algorithm}

\begin{algorithm}
\KwData{Input basis $\mathbf{L} = [\mathbf{l}_1, \mathbf{l}_2, ..., \mathbf{l}_r] \in \mathbb{R}^{m \times r}$, 
$\mathbf{d} \in \mathbb{R}^m$,
params $\lambda_1, \lambda_2, \{\eta\}_{1 \leq k \leq K}$, tol
}
\KwResult{projected samples $\mathbf{q} \in \mathbb{R}^r$ and $\mathbf{a} \in \mathbb{R}^m$
}   
$\mathbf{a} \leftarrow 0$ \;
\While {not converged}{
    Update the coefficient $\mathbf{q}$:\\
        $\quad \mathbf{q} \leftarrow (\mathbf{L}^\top\mathbf{L} + \lambda_1 \mathbf{I} + 2 \, \mathbf{L}^\top\mathbf{L} \sum_{k=1}^K \lambda_k)^{-1} \mathbf{L}^\top (\mathbf{d} - \mathbf{a} + 2\,\mathbf{L} \sum_{k=1}^K \lambda_k \mathbf{q}_{-T_k})$ \;
    Update the sparse component $\mathbf{a}$:\\
        $\quad \mathbf{a} \leftarrow \mathcal{S}_{\lambda_2} (\mathbf{d} - \mathbf{Lq})$ \;
    Converged if $\max \left\{ \frac{\Vert \mathbf{q}_{k+1} - \mathbf{q}_k \Vert}{\Vert \mathbf{d} \Vert}, \frac{\Vert \mathbf{a}_{k+1} - \mathbf{a}_k \Vert}{\Vert \mathbf{d} \Vert} \right\} < tol$
}
Return $\mathbf{q, a}$
 \caption{Sample projection}
 \label{algo:projection}
\end{algorithm}

\begin{algorithm}
\KwData{$\mathbf{L} = [\mathbf{l}_1, \mathbf{l}_2, ..., \mathbf{l}_r] \in \mathbb{R}^{m \times r}$, 
$\mathbf{B} = [\mathbf{b}_1, \mathbf{b}_2, ..., \mathbf{b}_r] \in \mathbb{R}^{r \times r}$,
$\mathbf{C} = [\mathbf{c}_1, \mathbf{c}_2, ..., \mathbf{c}_r] \in \mathbb{R}^{m \times r}$
}
\KwResult{Update basis $\mathbf{L}$}   
$\mathbf{\tilde{B}} \leftarrow \mathbf{B} + \lambda_1 \mathbf{I}$ \;
 \For{ $j = 1$ to $r$ }{
    $\mathbf{l}_j \leftarrow \frac{1}{\tilde{b}_{jj}} (\mathbf{c}_j - \mathbf{L}\mathbf{\tilde{b}_j}) + \mathbf{l}_j$
}
Return $\mathbf{L}$
 \caption{Fast basis update}
 \label{algo:basisUpdate}
\end{algorithm}

\subsection{Hyperparameters search}
There are numerous hyperparameters $\bm{\lambda} := [\lambda_1, \lambda_2, \{\eta_k\}]$ to be chosen and it is often unclear how to determine them. Without prior knowledge of the data, a cross-validation approach can be used. In practice, we select $J$ random subsets $\Omega_j \subset \Omega$ that we designate as missing entries. The cardinality of each subset is $\vert \Omega_j \vert = c \, \vert \Omega \vert$, where $c \in ]0, 1[$.
To obtain the optimal hyperparameters $\bm{\lambda}^*$, we use Bayesian Optimisation with Gaussian processes~\cite{scikit-learn} where the complex and expensive function to minimise is the average reconstruction error over the $J$ subsets $\Omega_j$: $\frac{1}{J} \sum_{j=1}^J \Vert \mathcal{P}_{\Omega_j}( D_{ij} - X(\bm{\lambda})_{ij}) \Vert_1$. We simply require that there are no values already missing from the $\Omega_j$ in order to evaluate the reconstruction.

Another parameter of prime importance is the estimation of the rank $r$. We have to provide the rank $r$ of $\mathbf{D}$ to initialise the matrices $\mathbf{L}$ and $\mathbf{Q}$. The rank $r$ has to be as small as possible in order to minimise the matrix sparsity and the low-rank error while keeping \textit{sufficient} information. A common practice is to look at the plot of the cumulative sum of the eigenvalues, \textit{e.g.} when one reaches $x\%$ of the nuclear norm (often at least 90\%).

\section{Experiments}
\label{sec:results}
We apply our framework to time series. First, we make up synthetic time series from sine functions. Second, we consider empirical data with a complex structure representing train ticketing validations. 

\subsection{Synthetic data}
As a first example of time series, we generate $J$ sine functions of $N$ points over [0, $2\pi N'$] with different frequencies $f_j$ and amplitude $a_j$, which we corrupt. More precisely, we delete data to create missing data (missing at random), we add anomalies that can reach twice the maximum amplitude of the initial function and finally, we add Gaussian noise. Time series are then added up; the period of the resulting time series equals $P$ = $N/N' \times$LCM$\{1/f_j\}_{j \in J}$ with LCM the Lowest Commun Multiple. We decide to reshape this time series into matrix of dimension $m \times n$, \textit{i.e.} $T_0 = m$ and $T_1 = P/m$. In this work, we set $N=10000, N'=10, m=100, f_1=1, f_2=3, f_3=1/2, a_1=a_2=1$ and $a_3=2$. Hence the resulting matrix is of dimension 100 $\times$ 100, $T_0 = 100$ and $T_1=20$. This procedure (corruption generation) is repeated 20 times and results reported are the average of these simulations. We make use of this simple case to observe the ability of the proposed frameworks to correctly disentangle the low-rank and sparse parts of the corrupted data. 

\begin{figure}[t]
    \centering
    \includegraphics[scale=0.4]{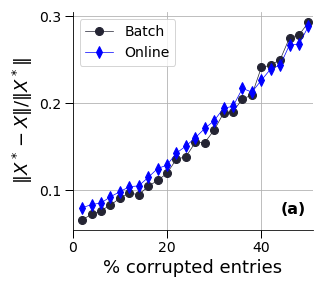}
    \includegraphics[scale=0.4]{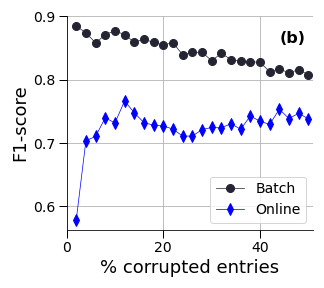}
    \includegraphics[scale=0.4]{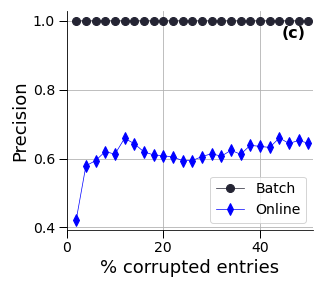}
    \caption{Comparison of two RPCA formulations: batch or online with additional temporal regularisations. (a) reconstruction errors; (b) F1-scores for the anomaly detection and (c) precision scores for the anomaly detection--with respect to the percentage of corrupted data.}
    \label{fig:results_sine}
\end{figure}

We start by evaluating the ability of the algorithms to reconstruct the underlying regular signal $\mathbf{X}$ by imputing the anomalies and missing data. The reconstruction errors are quantified by $\Vert \mathbf{X}^* - \mathbf{X} \Vert_F^2 / \Vert \mathbf{X}^* \Vert_F^2$. The batch and online versions offer similar reconstruction qualities regardless of the data corruption rate (FIG.~\ref{fig:results_sine}(a)). We also evaluate the performances in terms of anomaly detection. To do so, F1-scores and precision scores are computed. For these metrics, there is a clear superiority of the batch version over the online one (FIG.~\ref{fig:results_sine}(b) and (c)).

\subsection{Real data: ticketing validations time series}
\begin{figure}[!ht]
    \centering
    \includegraphics[scale=0.42]{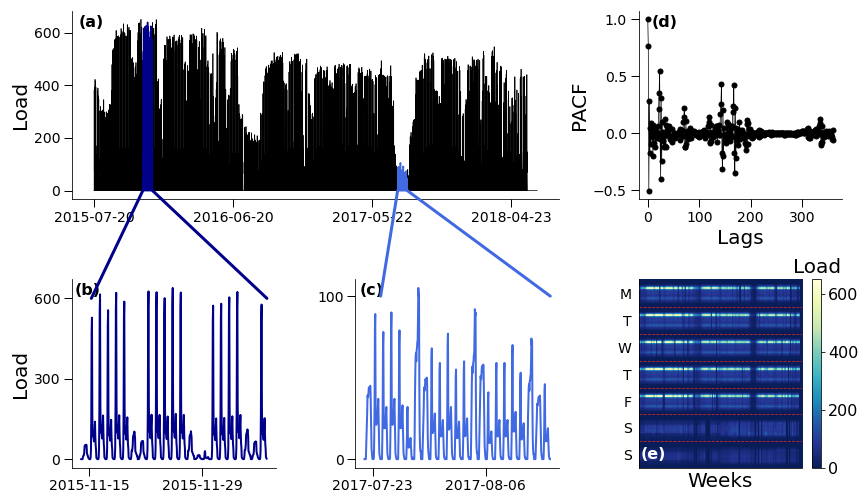}
    \caption{RPCA on a validation profile. (a) Entire time series of ticketing validations at the Athis-Mons station. (b) and (c) Focus on a subpart of the time series. (d) The time series displays a strong weekly seasonality, and to a lesser extent, a daily seasonality. (e) Matrix $\mathbf{D}$ associated to the time series in (a). Since the strong weekly seasonality, each column represents a week (dotted red lines separate the days Monday, Tuesday,$\dots$), and we can see that the weeks have similar patterns (high values in the morning of the working days).}
    \label{fig:sncf1}
\end{figure}

The methods are evaluated on Automatic Ticket Counting data which is the hourly data of transport ticket validations in stations of Ile-de-France\footnote{The data is available at \url{https://data.iledefrance-mobilites.fr/explore/dataset/histo-validations/information/}.}. The data cover the period January 2015 - June 2018 at an hourly granularity. They show strong weekly and, to a lesser extent, daily seasonality. The data also shows non stationary behaviours, such as local decreases in the number of validation (Fig.~\ref{fig:sncf1}(a)-(c)). In order to apply the RPCA algorithms, the time series is transformed into a matrix. Without any prior knowledge of the data, we can compute the time series partial autocorrelation function (PACF, Fig~\ref{fig:sncf1}(d)). One sees peaks corresponding to one-day and one-week correlations. For the RPCA algorithm to provide good results, a tip is to take $T_0$ as the lag giving the largest PACF value while having a low rank matrix $\mathbf{D}$. Here, the resulting matrix has $T_0=24\times7=168$ rows, \textit{i.e.} each week represents a column and we stack the successive 153 weeks vertically (see Fig.~\ref{fig:sncf1}(e)). We eventually get a matrix of dimension 168$\times$153.

\begin{figure}[!ht]
    \centering
    \includegraphics[scale=0.48]{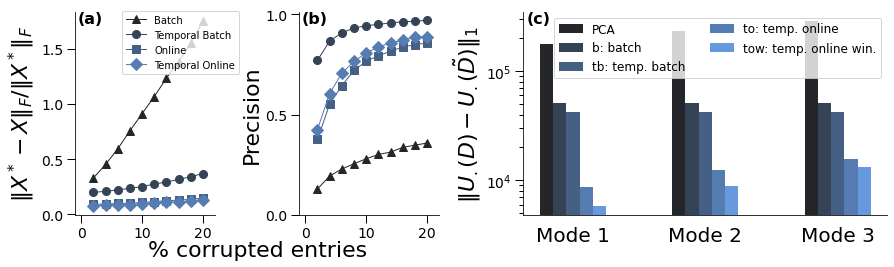}
    \caption{Comparison of four RPCA formulations on a validation profile: batch or online and with or without additional temporal regularisations ($T_1=1$, \textit{i.e.} similarity between consecutive weeks) (a) reconstruction errors and (b) precision scores for the anomaly detection; with respect to the percentage of corrupted data. The online versions offer the best reconstructions whilst the temporal ones are better at avoiding false positives. (c) RPCA filtering on modal decompositions: focus on the first modes computed on the resulting low-rank matrices.}
    \label{fig:sncf2}
\end{figure}

In order to quantify the performance of the proposed methods, we randomly corrupt some data and again inspect the reconstruction errors, F1-score and precision score as a function of the percentage of corrupted data. Temporal regularisations are computed with $T_1=1$, \textit{i.e.} similarity between consecutive weeks. Results are reported in Fig~\ref{fig:sncf2}. For this dataset, online versions offer better results than batch ones. Online RPCA are used with a moving window of size 50 (almost one year). A possible explanation for this behaviour may lie in the fact that the underlying subspace changes over time and thus the rank of matrix $\mathbf{L}$--which includes all observed samples--will increase over time. Retaining all past information may be more harmful than restricting oneself to the $n_{win}$ most recent ones. Indeed, as observed in Fig.~\ref{fig:sncf1}(e), it is possible that the subspace changes slightly at a yearly rate, giving the advantage to sliding window methods. 

We next explore the impact of RPCA filtering on modal decompositions. In particular, we add ten percent of corruptions to the observations $\mathbf{D}$ and obtain a corrupted matrix $\mathbf{\tilde{D}}$. We apply RPCA algorithms, \textit{e.g.} batch (b), temporal batch (tb), temporal online (to), and temporal online with window (tow) to $\mathbf{D}$ and $\mathbf{\tilde{D}}$ to get several matrices $\mathbf{X}$, \textit{i.e.} $\mathbf{X}_{b}(\mathbf{D}), \mathbf{X}_{o}(\mathbf{D}), ..., \mathbf{X}_{b}(\mathbf{\tilde{D}}), ...$ We then compute the PCA modes by the SVD decomposition on all these matrices with $\mathbf{M} = \mathbf{U \Sigma V^\top}$ the SVD decomposition of $\mathbf{M}$. We thus obtained several matrices $\mathbf{U}_{\text{PCA}}(\mathbf{D}), \mathbf{U}_{b}(\mathbf{D})$, ... Finally, we compute the difference between the first modes for each couple $(\mathbf{U}_{.}(\mathbf{D}), \mathbf{U}_{.}(\mathbf{\tilde{D}}))$. As illustrated in Fig~\ref{fig:sncf2}(c), the differences are smaller when applying an RPCA algorithm before calculating the decomposition; this show the interest of using RPCA to effectively recover coherent structures, when directly applying PCA fails. The smallest difference is for the temporal online with window method. 

\section{Discussion and conclusion}
\label{sec:discussion}

This work is motivated by real world data whose quality issues hinder their usage by machine learning algorithms. By representing a noisy and corrupted time series by an appropriate matrix, we have assume that the smooth signal can be recovered since it corresponds to a low-rank matrix, potentially perturbed by additive white noise, and some (sparse) anomalies. 
The temporal aspect of the data under scrutiny has allowed us to introduce a set of Toeplitz matrices $\mathbf{H_k}$ to improve some general RPCA frameworks. We have also proposed an online formulation that can be solved by a stochastic optimisation algorithm, offering a better computational complexity. The performed simulations have demonstrated the effectiveness of the proposed methods. Besides, we have proposed an end-to-end pipeline with a cross-validation strategy for the hyperparameters tuning. 

This work enlightens a new trade-off in the data reconstruction process. One would like to set the parameters that best lead to the recovery of the \textit{theoretical} low-dimensional subspace. In that case, we might have interest in selecting small dimension $r$. But at the same time,  we want to have an exact detection of anomalies in the observed data betting also sparse. The anomaly detection requires a good reconstruction, hence these two sparsity assumptions probably compete. 
We emphasise the important computational effort in the batch version required for obtaining hyperparameters that offer good and stable reconstruction performances. 
In this paper, our method is fully adaptive and requires little knowledge of the dataset: the Bayesian optimisation scheme is then well-adapted. 
Finally, RPCA frameworks are adapted for data with entries missing completely at random. One perspective could be to learn the distributions of these missing values to better impute them.

\bibliographystyle{unsrt}
\bibliography{arxiv}

\end{document}